\documentclass[11pt]{article}

\usepackage[%
]{acl}

\usepackage[T1]{fontenc}
\usepackage[utf8]{inputenc}
\usepackage{newtxtext}
\usepackage{newtxmath}
\usepackage{inconsolata}
\usepackage{microtype}

\usepackage{graphicx}
\usepackage{hyperref}
\usepackage{algorithm}
\usepackage{algpseudocode}
\usepackage{tcolorbox}
\usepackage{enumitem}
\usepackage{amsmath}
\usepackage{xfrac}
\usepackage{bm}
\usepackage{bbm}
\usepackage{booktabs}
\usepackage{tabularx}
\usepackage{multirow}
\usepackage{cleveref}
\usepackage{subcaption}
\usepackage{soul}
\crefname{algorithm}{algorithm}{algorithms}

\title{Identifying the Periodicity of Information in Natural Language}

\author{
 \textbf{Yulin Ou\textsuperscript{1}},
 \textbf{Yu Wang\textsuperscript{2}},
 \textbf{Yang Xu\textsuperscript{1}},
 \textbf{Hendrik Buschmeier\textsuperscript{2}}
\\
 \textsuperscript{1} CLCS Lab, Department of Computer Science and Engineering, \\Southern University of Science and Technology
 \\
 \textsuperscript{2} Digital Linguistics Lab, Department of Linguistics, Bielefeld University
\\
\\
 \small{
   \textbf{Correspondence:} \href{mailto:xuyang@sustech.edu.cn}{xuyang@sustech.edu.cn}
 }
}

\begin{document}
\maketitle

\begin{abstract}
Recent theoretical advancement of information density in natural language have raised the following question: To what degree does natural language exhibit \emph{periodicity} pattern in its encoded information? 
We address this question by introducing a new method called \texttt{AutoPeriod} of Surprisal (\texttt{APS}).  
\texttt{APS} adopts a canonical periodicity detection algorithm and is able to identify any significant periods that exist in the surprisal sequence of a single document. By applying the algorithm to a set of corpora, we have obtained the following empirical results: 
Firstly, a considerable proportion of human language demonstrates a strong pattern of periodicity in information. 
Secondly, new periods that are outside the distributions of typical structural units in text (e.g., sentence boundaries, elementary discourse units, etc.) are found and further confirmed via harmonic regression modeling. 
We conclude that the periodicity of information in language is a joint outcome from both structured factors and other driving factors that take effect at longer distances. 
The advantages of our periodicity detection method and its potentials in LLM-generation detection are further discussed. Our code is available as public repositories.\footnote{%
    Github: \url{https://github.com/CLCS-SUSTech/APS}\\
    Zenodo: \url{https://doi.org/10.5281/zenodo.19567886}}
\end{abstract}

\section{Introduction}

Understanding how information transmits in language is a long-standing research topic in linguistics \cite{shannon1948mathematical, shannon1949communication, shannon1951redundancy, bell2003effects}. One of the most influential assumptions is uniform information density \citep[UID;][]{aylett2004smooth, jaeger2010redundancy, jaeger2006speakers, meister-etal-2021-revisiting}, which states that information in language, measured by surprisal, distributes uniformly out of a need for efficient communication, as evidenced in its effect on, e.g., word duration in conversation and word abbreviation \cite{demberg-etal-2012-syntactic, MAHOWALD2013313}. Recent work further supplements this assumption with different empirical findings, such as the fluctuation of information with discourse boundaries \citep{xu2016entropy, xu2018information}; the thematic structures in dialogue that influence the change of information from speakers \citep{maes2022shared}.

These empirical findings give rise to an emerging research topic: the \emph{periodicity} of information in natural language. Periodicity is a bold hypothesis that goes a step further than the simple ``fluctuation'' of information, which is supported by recent evidence: the spectral features of text surprisal are distinguishable from white noise \citep{xu2017spectral, yang2023face}; spectra of surprisal are useful in machine-generation detection \citep{xu-etal-2024-detecting, face2-emnlp-2025}; surprisal variation can be modeled by harmonic regression, which confirms periodic patterns across various linguistic structures \citep{tsipidi-etal-2025-harmonic}. 

However, some key questions about this topic remain unanswered. First of all, how can we find \textbf{direct evidence of the periodicity} of information in language? 
The Fourier-analysis-based spectral methods proposed in \citet{yang2023face} and \citet{face2-emnlp-2025} provide useful insights, but can only serve as indirect evidence of periodicity, because the spectrum in frequency-domain was not transformed back to explicit periods in time-domain. 
On the other hand, the harmonic regression method proposed by \citet{tsipidi-etal-2025-harmonic}, provides a way to validate periodicity in information, given ``candidate'' periods selected from certain linguistic structures (e.g., sentences, paragraphs, etc.). 
Therefore, it potentially filters out periods that cannot be attributed to any pre-defined linguistic units. And its reliance on regression analysis prevents it from determining if a text segment is truly periodic. 
This leads to our second question: can we \textbf{detect periodicity at the document level} more precisely and with higher quality?
That is, given a piece of text, we want to find all periods that can pass a certain detection threshold with some clearly defined confidence levels, or otherwise to determine its non-existence. As detection occurs at the document level, such a method would allow us to explore whether periodicity correlates with other theoretically interesting factors, such as text genre, authorship, and the distinction between human-written texts and those generated by large language models (LLMs). 
 
In this study we address the above-mentioned questions by proposing a periodicity detection algorithm named \texttt{AutoPeriod} of Surprisal (\texttt{APS}), which is able to directly identify any existing periods in an input sequence of surprisal at certain confidence levels (see \Cref{sec:period_detect}). 
We apply this algorithm to common linguistic corpora and have obtained positive evidence for periodicity of information (\Cref{sec:results_identified_periods}). 
The validity of our results is verified with an alternative method (\Cref{sec:results_comparison_hr}). The potential merits of periodicity detection in distinguishing LLM-generations (\Cref{sec:results_human_vs_generated}).

\section{Background}

\subsection{UID and the Natural Fluctuation of Information}

The UID theory describes that a general trend of how the information of linguistic units distributes within the course of communication is to stay relatively uniform or constant. 
However, token-level uniformity/constancy is an asymptotic property that would only manifest when observation is conducted over a sufficiently long sequence, during which the natural fluctuation of information would inevitably occur. 
For example, \citet{genzel2002entropy, genzel2003variation} observed the increasing surprisal near the beginning of paragraphs and the sudden drop near the end. It can be imagined that if multiple paragraphs of similar lengths are organized together as a chapter, then an overall regular fluctuation (or even \emph{periodicity}) will obviously appear.

Building on the coupling between uniformity and fluctuation illustrated above, recent works have attempted to model the latter more formally. 
A recurring convergence pattern of surprisal is found within topic boundaries in dialogue \citep{xu2016entropy, xu2018information}. It is also found that the uniformity trend is only salient in certain contextual units \citep{giulianelli2021information} and not always observed \citep{giulianelli2021analysing}. 
More recently, surprisal fluctuation over discourse has been modeled with hierarchically-structured Bayesian regression \citep{tsipidi-etal-2024-surprise} and harmonic regression \citep{tsipidi-etal-2025-harmonic}.

\subsection{Why Study the Periodicity of Surprisal}

Given that surprisal can describe how linguistic information changes dynamically in human language, its potential uses have been explored, such as (i) measuring creative use of language such as metaphor and humor \cite{bunescu-uduehi-2022-distribution, xie-etal-2021-uncertainty}; (ii) evaluating the quality of generated text \cite{kharkwal2014surprisal,yang2023face,face2-emnlp-2025}; (iii) and, recently, detecting LLM-generated text \cite{xu-etal-2024-detecting}. 

A further study on the periodicity of surprisal will largely facilitate the above-listed research directions. While UID theory and its variants can make asymptotic predictions about the general trends of information in language, using periodicity as a lens would provide more fine-grained insights into the instantaneous dynamics of information, which better meets the purpose of studying the token-by-token generation process. 
What's missing in the picture is a method and toolkit that can effectively extract periodicity as a clear feature from text  -- current regression-based methods \citep{tsipidi-etal-2025-harmonic} provide ways to validate but not to comprehensively identify periodicity; Fourier-based methods \citep{xu-etal-2024-detecting} lack direct interpretation. Therefore, the goal of this study is to fill this gap. 

\section{Periodicity Detection}
\label{sec:period_detect}

The method proposed in this study for detecting periodicity in information is based on the \texttt{AutoPeriod} algorithm \citep{autoperiod}. Originally designed for detecting periodicity in economics and social time series data, this algorithm demonstrates reliable performance in identifying periodic patterns.
We adaptively modify and re-implement this algorithm -- tailoring its parameters to the characteristics of surprisal -- and release it as a standalone toolkit called \texttt{AutoPeriod} of Surprisal (\texttt{APS}). 
 
The algorithm runs two sub-procedures: 
\begin{enumerate}[label=\arabic*), ref=\arabic*, leftmargin=1.2em]
    \item \textsc{GetPeriodHints}: Gets the periodogram spectrum of an input sequence and extracts the candidate periods called \emph{period hints}, whose power values are above a certain threshold obtained from random permutations. 
    \item \textsc{ACFFiltering}: Filters the period hints using the auto-correlation function (ACF), and only keep those that lie on the hills of the ACF curves.  These are considered as the valid periods.
\end{enumerate}

We first introduce the mathematical prerequisites for the Fourier transform and the periodogram in \Cref{sec:prereqs}, as these foundational concepts are essential for understanding the \texttt{AutoPeriod} procedure. 

\subsection{Prerequisites: Fourier Transform and Periodogram}
\label{sec:prereqs}

Given a piece of text that contains $N$ tokens, we first estimate the surprisal of each token by computing the negative log probability $x_n=-\log p(t_n \mid t_{<n})$, where $p$ is the probability of token $t_n$ predicted by a language model based on its preceding context $t_{<n}$.
The resulting sequence of surprisal values $\mathbf{x}:=x_0,\dots,x_{N-1}$ will be the input to the discrete Fourier transform (DFT), defined as:
\begin{equation*} 
    X_k = \sum_{n=0}^{N-1}x_n e^{-i2\pi\frac{k}{N} n}\label{eq:DFT}
\end{equation*} 
where $X_k$ is the cross correlation of $\mathbf{x_n}$ and a complex sinusoid, $e^{-i2\pi\frac{k}{N}n}=\cos(2\pi\frac{k}{N}n)-i\sin(2\pi\frac{k}{N}n)$ ($n=0,\dots,N-1$). 

The norm $\lVert X_k \rVert = \sqrt{\textit{re}(X_k)^2+\textit{im}(X_k)^2}$ measures to what degrees the original input can be reconstructed from the periodic signal at frequency $2\pi\frac{k}{N}$. 
The raw DFT output $\mathbf{X}:=X_0,\dots,X_{N-1}$ has the property of being \emph{even symmetric}, that is, $\lVert X_k \rVert = \lVert X_{N-k} \rVert $, $\forall k\in 0,\dots,N-1$. Therefore, it is sufficient to use the first half of $\mathbf{X_k}$, $k\in 0,\dots, \lceil \frac{N-1}{2} \rceil$ to encode the full spectral information. 
This truncated magnitude sequence is also known as \textbf{periodogram},\footnote{%
    The periodogram algorithm we introduced here follows the design in \citet{autoperiod}. In our experiments, we tried variants of periodogram, and selected the Lomb-Scargle periodogram \citep{lomb, scargle} in order to enhance the performance of \texttt{APS} in low-frequency range (i.e., long periods).}
a widely-used technique in signal processing \citep{schuster1898investigation, oppenheim1999discrete}. 
We denote the periodogram as $\bm{\mathcal{P}}=\{\lVert X_k\rVert\}$ ($k\in 0,\dots, \lceil \frac{N-1}{2} \rceil$).

\subsection{Step 1: Extract Period Hints From the Periodogram}
\label{sec:autoperiod1_hints}

The \textsc{GetPeriodHints} procedure is described in \Cref{alg:period_hints}. Within this step, the surprisal sequence is first permuted for $m$ times, and the maximum spectrum powers from these permuted sequences are recorded in $\textit{maxPower}$. The $\textit{percentile}$-th largest value in $\textit{maxPower}$ is used as the power threshold $\mathcal{P}_{\text{threshold}}$ to find those frequencies/periods whose powers are above this threshold. Therefore, the period hints returned are the most significant ones with a $\textit{percentile}/100$ confidence level. 

\begin{algorithm}
\caption{First step of \textsc{AutoPeriod}: Extract period hints.}\label{alg:period_hints}
\begin{algorithmic}[1]
\Function{getPeriodHints}{$\mathbf{x}$, $m$, $\textit{percentile}$}
    \State $N \gets \mathbf{x}.\textit{length}$ \Comment{length of input sequence}
    \State $m \gets 100$ \Comment{number of permutations}
    \State $\textit{percentile} \gets 99$
    \State $\textit{maxPower} \gets \{\}$ \Comment{empty set}
    \State $\textit{periodHints} \gets \{\}$
    \For{$i = 1$ to $m$}
        \State $\mathbf{x^{\text{rand}}} \gets \textsc{Permute}(\mathbf{x})$
        \State $\bm{\mathcal{P}^{\text{rand}}} \gets \textsc{getPeriodogram}(\mathbf{x})$
        \State $\textit{maxPower}.\textsc{add}(\max(\bm{\mathcal{P}^{\text{rand}}}))$
    \EndFor
    \State $\textit{maxPower}.\textsc{sort}()$ \Comment{ascending}
    \State $P_{\text{threshold}} \gets \textit{maxPower}[m \cdot (\frac{\textit{percentile}}{100})]$
    \State $\bm{\mathcal{P}} \gets \textsc{getPeriodogram}(\mathbf{x})$
    \For{$k = 1$ to $\mathbf{\mathcal{P}}.\textit{length}$}
        \If{$\bm{\mathcal{P}}[k] > P_{\text{threshold}}$}
            \State $\textit{periodHints}.\textsc{add}(\frac{N}{k})$ 
        \EndIf
    \EndFor
    \State \Return $\textit{periodHints}$
\EndFunction
\end{algorithmic}
\end{algorithm}

\begin{figure*}
    \centering
    \includegraphics[width=0.95\linewidth]{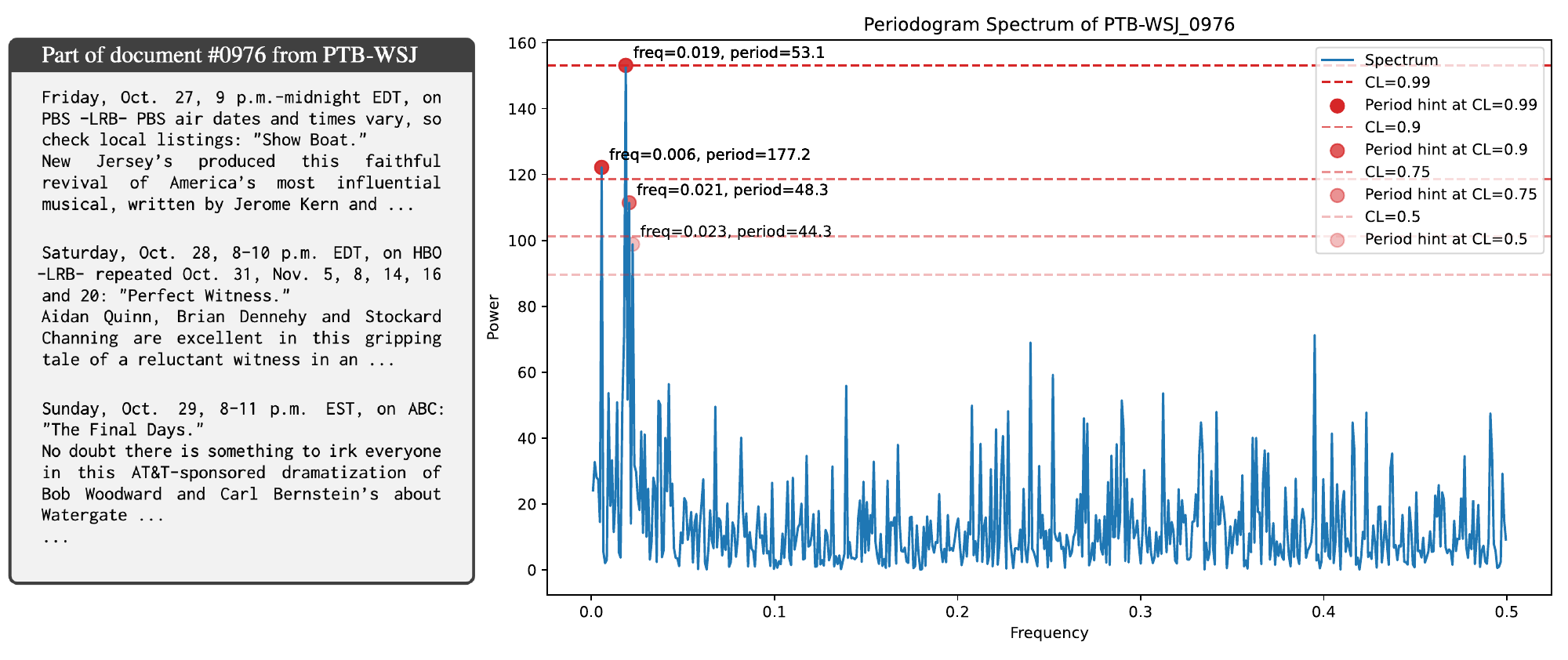}
    \caption{Periodogram spectrum obtained from the surprisal sequence of document \#0976 in PTB-WSJ corpus. Right: Red dots indicate the period hints returned by \Cref{alg:period_hints}. Dashed lines are the power thresholds used at different confidence levels.}
    \label{fig:period_hints}
\end{figure*}

Here we use document \#0976 from PTB-WSJ as a complete example of analysis. The texts are snippets of TV listings with highly structured format of content -- with weekdays followed by channel names and short summaries (see \Cref{fig:period_hints}, left). 
The result of running \textsc{GetPeriodHints} on this document is shown in \Cref{fig:period_hints} (right). It can be seen that when the confidence level (CL) decreases from $.99$ to $.50$, more period hints are extracted. The most confident ($\text{CL}=.99$) period hint occurs at the frequency peak $f\approx.019$, and the corresponding period $T=\frac{1}{f}\approx53.1$ (tokens), suggesting that a period of surprisal occurs roughly every 53 tokens. 
$\text{CL}=.90$ is used throughout this study. 
Before a period hint is validated in the next step, we deem it as quasi-periodic.

\subsection{Step 2: Period Validation via ACF Filtering}
\label{sec:autoperiod2_acf}

\begin{figure}[ht]
    \centering
    \includegraphics[width=\linewidth]{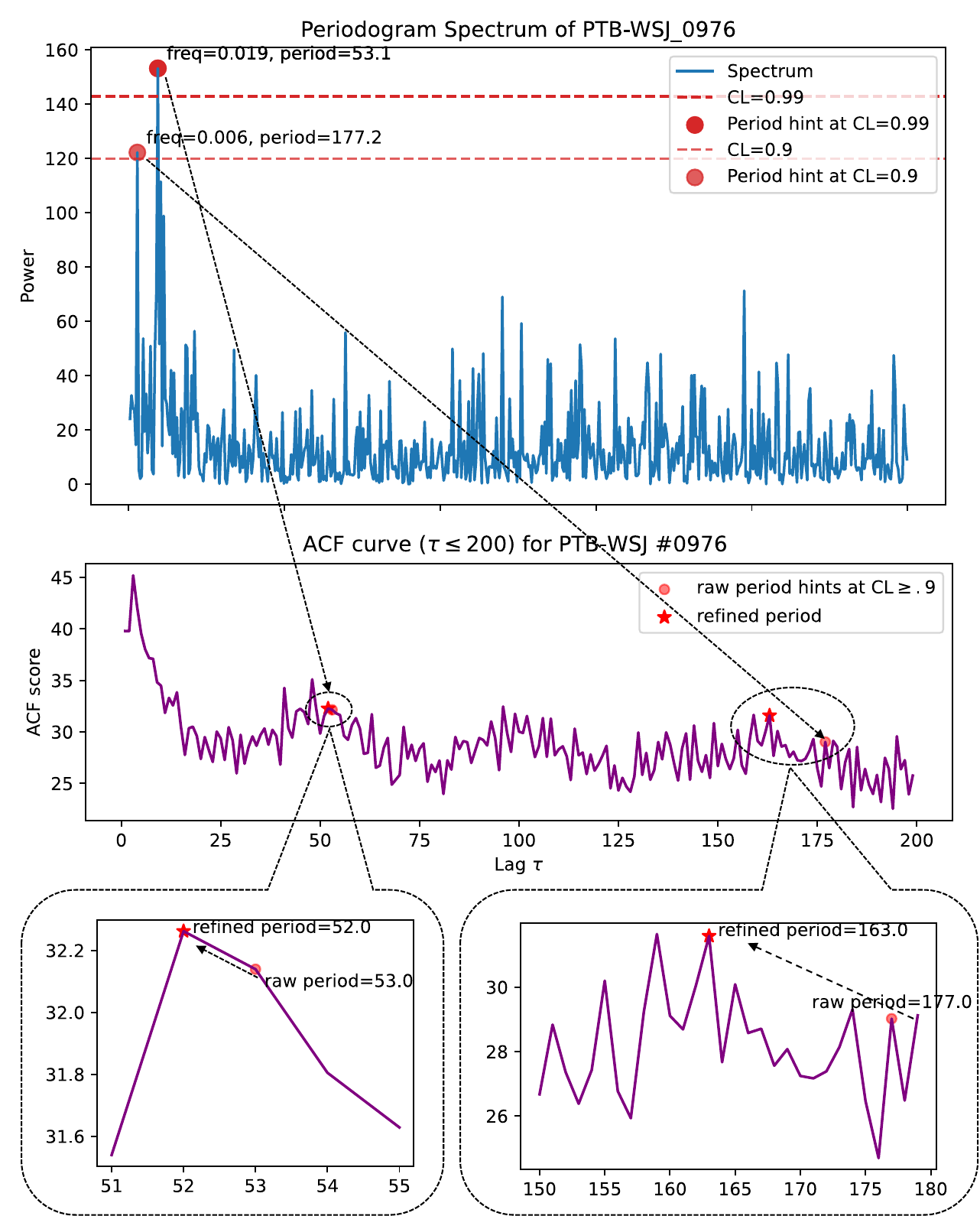}
    \caption{ACF filtering for surprisal sequence of document \#0976 from the PTB-WSJ corpus. Period Hints (red dots; top) are mapped to the raw periods in the ACF curve (red dots; middle). ACF filtering refines and returns the final periods (red stars; bottom).}
    \label{fig:period_acf}
\end{figure}

The brief version of \textsc{ACFFiltering} procedure at Step 2 is described in \Cref{alg:period_acf}. This involves applying a fine-grained validation to the period \emph{hints} found in Step 1, which allows spurious periods and false alarms to be filtered out and the \emph{valid} periods to be identified following further refinement. This filtering step is initially designed to increase the detection precision for periodicity in generic time series, and briefly, the motivation is that a \emph{dominant} period $\tau$ of a time series $x$ (of length $N$) should have a relatively large value of the autocorrelation function (ACF):
\begin{equation*}
    \textrm{ACF}(\tau)=\frac{1}{N}\sum_{n=0}^N x(\tau)\cdot x(n+\tau),
\end{equation*}
which measures how similar the sequence is to its previous values for different $\tau$ lags. 
If a period hint $\tau$ from the periodogram lies on a hill of the ACF curve, then we are more confident to consider it as a valid period, otherwise it is likely to be a false alarm. 
This validation is done to each period hint $\tau=\frac{N}{k}$, by examining the nearby points on the ACF curve within a window of size $W_{N/k}=[a,b]$. 
If $\tau$ does not lie on a hill, this period hint is discarded, otherwise a further \emph{refinement} is done by searching within the nearby region and finding the closest local maximum of the ACF scores. 
As can be seen in \Cref{fig:period_acf} (bottom), the raw/unrefined period hints $\tau'_1=53.2$ and $\tau'_2=177.2$ are refined to the final integer periods $\tau_1=52$ and $\tau_2=163$, respectively.

\begin{algorithm}
\caption{ACF Filtering (brief version)}\label{alg:period_acf}
\begin{algorithmic}[1]
\Function{ACFFiltering}{$\mathbf{x}$, $\textit{periodHints}$}
    \State $N \gets \mathbf{x}.\textit{length}$
    \State $\textit{ACF} \gets \textsc{getACF}(\mathbf{x})$
    \State $M \gets \textit{periodHints}.\textit{length}$
    \State $\textit{refinedPeriods} \gets \{\}$ \Comment{empty set}
    \For{$i = 1$ to $M$}
        \State $\tau \gets N/k \gets \textit{periodHints}[i]$
        \State $\tau' \gets$ local maximum within a range
        \If{$\tau'$ is on hill of ACF}
            \State $\textit{refinedPeriods}.\textsc{add}(\tau')$
        \EndIf
    \EndFor
    \State \Return \textit{refinedPeriods}
\EndFunction
\end{algorithmic}
\end{algorithm}

\section{Corpora and Language Models}
\label{sec:corpora}

This study utilizes several text corpora. Penn Treebank \citep[PTB;][]{marcus1993building} and Chinese Treebank
\citep[CTB;][]{xue2013chinese} are adopted as representatives of human-written texts in English and Chinese, respectively. Specifically, the Wall Street Journal (WSJ) and Brown portions are used from PTB.
Moreover, to explore the relationship between the global periodicity detected by \texttt{APS} and the local linguistic structures focused on by \citet{tsipidi-etal-2025-harmonic}, we incorporate the Rhetorical Structure Theory (RST) based Discourse Treebank \citep{RSTcorpus} for English and the GCDT corpus \citep{peng_gcdt_2022} for Chinese. These corpora provide discourse-level annotations, enabling a comparative analysis. The detection results of the corpora mentioned above are mainly discussed in \Cref{sec:results_identified_periods} and \Cref{sec:results_comparison_hr}.

To investigate the periodicity differences between human-written and LLM-generated texts, we use two datasets: FACE collected by \citet{face2-emnlp-2025} and EvoBench by \citet{yu-etal-2025-evobench}. Our analysis focuses on the English news and Chinese news portions of FACE, and the XSum portion of EvoBench. 
For each portion, we select a subset generated by the largest models (LLaMA3-70B for English FACE,  Qwen2-72B for Chinese FACE, GPT-4o for EvoBench).
The specific subsets used are detailed in \Cref{tab:human_vs_generated}, with results presented in \Cref{sec:results_human_vs_generated}.

To extract surprisal sequences from the corpora, LLMs are used to encode the input texts. For English texts, LLaMA3-8B \citep{touvron2023llama} is adopted. For Chinese texts, Qwen2-7B \citep{yang2024qwen2} is adopted. Specifically, to ensure comparability with the experiments of \citet{tsipidi-etal-2025-harmonic}, we use Yarn-LLaMA2-7B \citep{peng2024yarn} for RST Discourse Treebank.

\section{Result: Identified Periods of Information}\label{sec:results_identified_periods}

\subsection{Proportions of Periodic Documents}\label{sec:proportions_periodic_docs}

\begin{figure}
    \centering    
    \vspace{2mm}
    \includegraphics[width=0.7\linewidth]{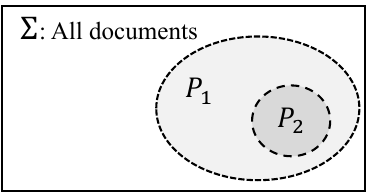}
    \caption{Partitioning a corpus based on the periodicity detection outcomes from \texttt{APS}.}
    \label{fig:dataset_partition}
\end{figure}

Applying \texttt{APS} on a text corpus, indicated by $\Sigma$, would partition it into three subsets: $P_1$, the set of documents that contain at least one period hint (not necessarily a valid period); $P_2$, the set of documents that contain at least one valid period; and $\Sigma-P_1$, the remaining set containing no period hints or valid periods (see \Cref{fig:dataset_partition}).

Across the corpora examined, we consistently find that \textbf{a non-negligible proportion of texts exhibit strong periodicity of information}. 
\Cref{tab:hints_valid_ratios} shows the counts of documents $|\Sigma|$, $|P_1|$ and $|P_2|$, and the proportions $\sfrac{|P_1|}{|\Sigma|}$, $\sfrac{|P_2|}{|P_1|}$ and $\sfrac{|P_2|}{|\Sigma|}$. 
$P_2$ are the most strictly periodic documents, whose proportion
$\sfrac{|P_2|}{|\Sigma|}$ reaches above 5\% for all corpora. 
$P_1$ is the quasi-periodic set, and its odds $\sfrac{|P_1|}{|\Sigma|}$ is significantly higher, reaching about $15\%$ on average. 
The relatively high ratio of $\sfrac{|P_2|}{|P_1|}$ (about $66\%$) indicates that once a hint is detected by \textsc{GetPeriodHints} (Step~1), it has a large chance to pass \textsc{ACFFiltering} (Step~2). 

All three ratios are higher in Chinese corpora than in English. The exceptionally high values of GCDT against the rest also suggests that text genre is an important factor affecting periodicity -- GCDT contains five genres of formal texts (academic articles, biographies (bio), interview conversations, news, and how-to guides (whow)), while WSJ, Brown, and CTB are solely collected from news articles. 
We will leave the comprehensive examination on how text genre (such as the dimensions of verse vs. prose, oral vs. written, etc.) influences the periodicity of information as future work. 

\subsection{Comparing \texttt{APS} Periods With Linguistic Structural Units
}\label{sec:comparing_periods_with_units}

\begin{figure}
    \centering
    \includegraphics[width=\linewidth]{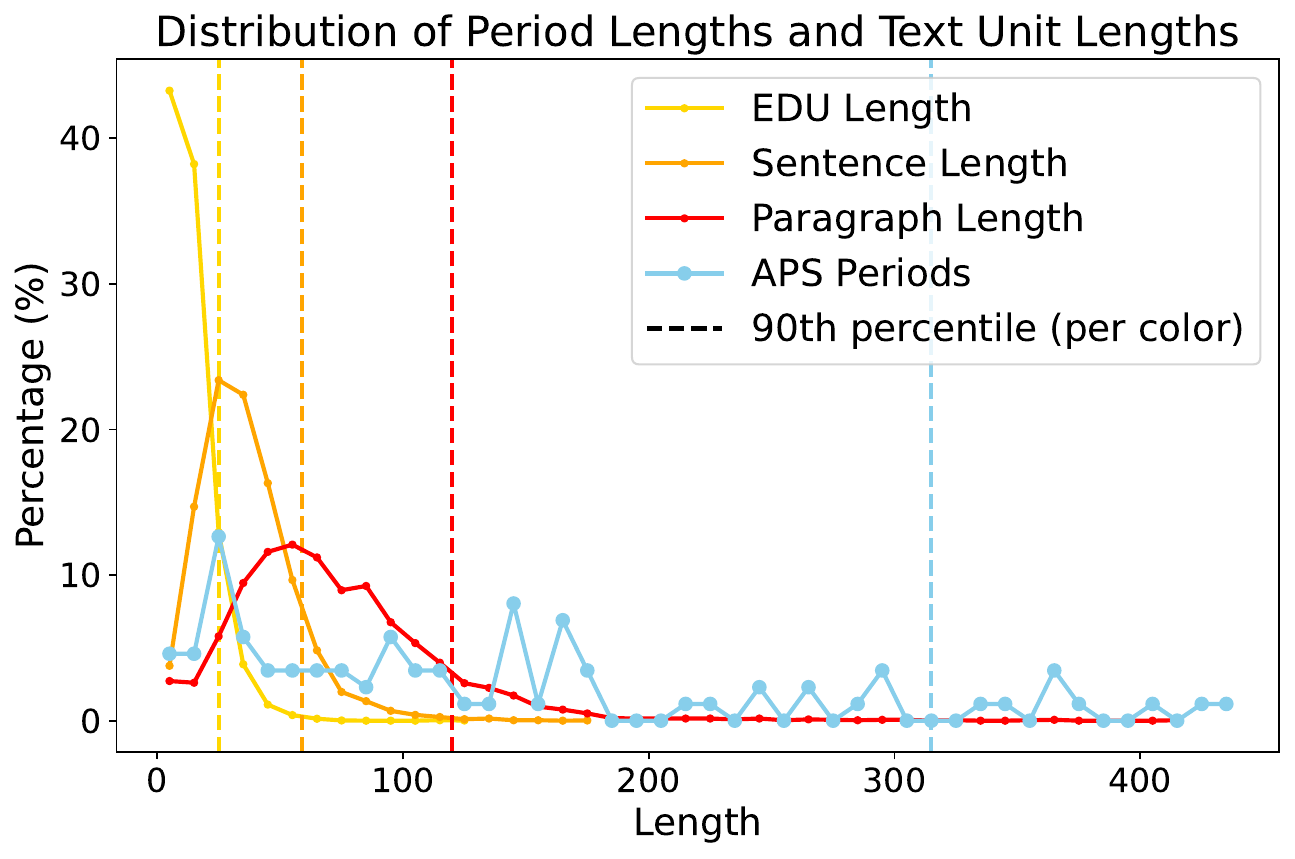}
    \caption{Distribution of valid periods identified by \texttt{APS} on Discourse Treebank, and the length distributions of structural units (EDUs, sentences, paragraphs). The dashed lines indicate the 90th percentiles of the corresponding distributions.}
    \label{fig:periods_dist_and_unit_mean}
\end{figure}

To assess the relationship between the global periodicity detected by \texttt{APS} and the local linguistic structures focused on by \citet{tsipidi-etal-2025-harmonic}, we conducted a quantitative comparison between length distribution of the valid periods detected by \texttt{APS} and those of three structural units: elementary discourse units (EDUs) annotated in the RST Discourse Treebank, sentences, and paragraphs. The distribution of the detected periods is presented in \Cref{fig:periods_dist_and_unit_mean}, 
along with the length distributions of the structural units.
The most prominent peak in the period distribution is observed around 25 tokens, which aligns closely with the peaks of length distributions from the three structural units.
This alignment suggests that a subset of the detected periodic patterns can indeed be attributed to these fundamental structural units.

However, a substantial proportion of the detected periods fall outside the typical length distributions of the three considered structural units.
Notably, the \emph{paragraph} (the longest among the three) has $3.63\%$ of its instances exceeding 150 tokens, and only $1.11\%$ exceeding 200 tokens. In contrast, there are $33.33\% $of the valid periods detected by \texttt{APS} that exceed 150 tokens, and $21.84\%$ that exceed 200 tokens. The significant portion of long-period patterns identified by \texttt{APS} cannot be accounted for by the structural units.
This clear mismatch indicates that \textbf{the global periodicity of information identified by \texttt{APS} is not fully explicable by the succession of local discourse structures alone}. Instead, it suggests the presence of additional, yet-to-be-identified, factors contributing to the long-periodicity of information in text.

\begin{table*}[!ht]
\centering
\caption{Counts of document with non-empty period hints ($|P_1|$) and valid periods ($|P_2|$) identified by \texttt{APS}, along with corresponding ratios relative to the total document counts ($|\Sigma|$) across four corpora. The last two rows are the mean and median lengths of the hints/valid periods identified in $P_1$ and $P_2$, respectively.}

\begin{tabularx}{\linewidth}{lXXXXX}
\toprule
          & $\text{WSJ}_{\text{en}}$       & $\text{Brown}_{\text{en}}$     & $\text{GCDT}_{\text{zh}}$      & $\text{CTB}_{\text{zh}}$       & Avg.      \\
\midrule
$|\Sigma|$           & 2499      & 500       & 50        & 2773      & ---         \\
$|P_1|$            & 221       & 52        & 15        & 304       & ---         \\
$|P_2|$           & 131       & 25        & 13        & 212       & ---         \\
\midrule
$\sfrac{|P_1|}{|\Sigma|}$     & 8.84\%    & 10.40\%   & 30.00\%   & 10.96\%   & 15.05\%   \\
$\sfrac{|P_2|}{|P_1|}$      & 59.28\%   & 48.08\%   & 86.67\%   & 69.74\%   & 65.94\%   \\
$\sfrac{|P_2|}{|\Sigma|}$     & 5.24\%    & 5.00\%    & 26.00\%   & 7.65\%    & 10.97\%   \\
\midrule
$P_1$: mean (median) & 127.68 (100) & 111.37 (28) & 113.76 (81) & 97.53 (51) & 112.36 (74) \\
$P_2$: mean (median) & 135.65 (108) & 144.57 (84) & 110.53 (56) & 89.69 (34) & 113.14 (70) \\
\bottomrule
\end{tabularx}
\label{tab:hints_valid_ratios}
\vspace{-0.5mm}
\end{table*}

\section{Result: Validity-Check with Harmonic Regression Modeling}
\label{sec:results_comparison_hr}

In this section, we use an alternative method to carry out a validity-check on the detection results from \texttt{APS}. \citet{tsipidi-etal-2025-harmonic} formulates periodicity of information as the effect of harmonic components in predicting the surprisal of a token $w_t$ using harmonic regression (HR) models. In their models, periods are defined as the lengths of structural units (EDUs, sentences, and paragraphs). The prediction task is formulated as: 
\begin{equation}
\label{eq:hr_pred}
    s(w_t) \sim \text{baseline} + \text{HR}(U_t)
\end{equation}
in which ``baseline'' indicates the same basic predictors (e.g., relative positions of $w_t$, etc.) used in \citet{tsipidi-etal-2025-harmonic}, and the $\text{HR}(U_t)$ term is the linear combination of sine and cosine components corresponding to a certain unit length $U_t$: 
\begin{equation}
\label{eq:hr_term}
\begin{aligned}
    \text{HR}(U_t) = \beta_0 + \sum_{k=1}^{K} & \Bigg( \beta_{1,k}\cdot \sin\left(\frac{k2\pi t}{U_t}\right) \\
    &+ \beta_{2,k}\cdot \cos\left(\frac{k2\pi t}{U_t}\right) \Bigg)
\end{aligned}
\end{equation}
Significant coefficients in the form of amplitude  $A_k=\sqrt{\beta_{1,k}^2+\beta_{2,k}^2}$ from the $\text{HR}(U_t)$ terms would indicate that certain degrees of periodicity exist within the scope of $U_t$. 
Based on this paradigm, we proceed with two inquires:
\begin{enumerate}[leftmargin=1.2em]
    \item Does using the hints and periods from \texttt{APS} (our method) as the scaling factor $U_t$ result in significant HR modeling outcomes?
    \item Do the documents that return positive in \texttt{APS} detection ($P_2$) result in better overall performance in HR modeling?
\end{enumerate}

\subsection{Hints/Valid Periods Directly Used as Scaler in HR}

We replace $U_t$ in \Cref{eq:hr_pred} with the \emph{hints} and valid \emph{periods} returned by \texttt{APS}, and then examine the corresponding HR modeling outcomes. If there are multiple hints or periods in one document, then the longest one is selected. The modeling formulas then become: 
\begin{align*}
    m_{\text{hints}}: s(w_t) &\sim \text{baseline}+\text{HR}(\textit{hints})\\
    m_{\text{periods}}: s(w_t) &\sim \text{baseline}+\text{HR}(\textit{periods})
\end{align*}
The models $m_{\text{hints}}$ and $m_{\text{valid period}}$ are fit against the $P_1$ and $P_2$ portions of WSJ, respectively (to be directly comparable to \citeposs{tsipidi-etal-2025-harmonic} results, only WSJ is used). 
The coefficients of the harmonic terms $\text{HR}(\textit{hints})$ and $\text{HR}(\textit{periods})$ with the top-3 largest amplitude values are reported in \Cref{tab:wsj_coef_P1,tab:wsj_coef_P2}. The results show that both terms are indeed significant predictors in HR modeling, which confirms the periodicity of hints and valid periods that \texttt{APS} returns. 

There is no significant differences between the $A_k$ values of hints and valid periods. This suggests that, although the latter is stronger than the former by the standard of passing the periodicity threshold, their modeling effects on predicting local surprisal are almost equivalent. 
We also find that the $A_k$ values from hints and valid periods are smaller than those from EDU, the best predictor reported in \citet{tsipidi-etal-2025-harmonic}. This is to be expected, as hints/valid periods are fixed values within a document and therefore less dependent on the position of tokens. This naturally makes them less informative predictors. 
However, this does not affect their periodicity. Rather, it strengthens our conclusion that \texttt{APS} can detect global periodic patterns that differ from local units. 

\begin{table}
\centering
\caption{Significance test of the top-3 harmonic components with the largest amplitudes in HR models using \textbf{\texttt{APS} \emph{periods}} as scaling factors on WSJ ($P_2$ portion).}
\label{tab:wsj_coef_P2}
\begin{tabularx}{\linewidth}{lXlXX}
\toprule
$k$ & $A_k$ & $\beta$ & coef & $P > |t|$ \\
\midrule
\multirow{2}{*}{1} & \multirow{2}{*}{0.2022}
 & $\beta_{1,1}$ & 0.1213 & 0.000 \\
 & & $\beta_{2,1}$ & 0.1571 & 0.000 \\
\midrule
\multirow{2}{*}{3} & \multirow{2}{*}{0.0652}
 & $\beta_{1,3}$ & 0.0524 & 0.000 \\
 & & $\beta_{2,3}$ & 0.0405 & 0.000 \\
\midrule
\multirow{2}{*}{4} & \multirow{2}{*}{0.0467}
 & $\beta_{1,4}$ & 0.0367 & 0.002 \\
 & & $\beta_{2,4}$ & 0.0289 & 0.010 \\
\bottomrule
\end{tabularx}
\end{table}

\begin{table}
\centering
\caption{Significance test of the top-3 harmonic components with the largest amplitudes in HR models using \textbf{\texttt{APS} \emph{hints}} as scaling factors on WSJ ($P_1$ portion).}
\label{tab:wsj_coef_P1}
\begin{tabularx}{\linewidth}{lXlXX}
\toprule
$k$ & $A_k$ & $\beta$ & coef & $P > |t|$ \\
\midrule
\multirow{2}{*}{1} & \multirow{2}{*}{0.1772}
 & $\beta_{1,1}$ & 0.0736 & 0.000 \\
 & & $\beta_{2,1}$ & 0.1612 & 0.000 \\
\midrule
\multirow{2}{*}{5} & \multirow{2}{*}{0.0731}
 & $\beta_{1,5}$ & 0.0503 & 0.000 \\
 & & $\beta_{2,5}$ & -0.0531 & 0.000 \\
\midrule
\multirow{2}{*}{7} & \multirow{2}{*}{0.0575}
 & $\beta_{1,7}$ & 0.0036 & 0.723 \\
 & & $\beta_{2,7}$ & -0.0574 & 0.000 \\
\bottomrule
\end{tabularx}
\end{table}

\subsection{Filtering Data Using Hints/Valid Periods Improves HR Performance}

To test whether \texttt{APS} can filter out data with better overall periodicity, we fit HR models on four different portions, $\{P_2, P_1, \Sigma,\Sigma-P_1\}$ of RST and GCDT corpora, using two structural units (EDUs, sentences) along with whole texts as different scaling factors. 
We assume that the resulting mean squared errors (MSE) should rank in a descending order, i.e., $P_2 < P_1 < \Sigma < \Sigma - P_1$, as $P_2$ is the most periodic portion according to \texttt{APS}, while $\Sigma-P_1$ is the least. 

The results in \Cref{tab:wsjhr,tab:gcdt} confirm our assumption. The rankings remain consistent across different scaling factors and even so in baseline models (without HR terms). This indicates that the documents identified by \texttt{APS} as containing strong periodicity of information are more predictable in terms of token surprisal. Scaling by structural units further improves the performance on the filtered portions, and EDU scaling is the most effective one, which is consistent with \citeposs{tsipidi-etal-2025-harmonic} results. From this perspective, the global periodicity identified by \texttt{APS} and the local periodicity modeled by HR using structural units are compatible. 
It is possible that the periodic patterns of discourse structures contribute to the overall periodicity of information in text.

\begin{table}
\centering
\caption{MSE of HR modeling on \textbf{WSJ} using different scaling factors and different data portions.}
\label{tab:wsjhr}
\small
\begin{tabularx}{\linewidth}{llllX}
\toprule
 & Baseline & EDU & Sentence & Whole Text \\
\midrule
$P_2$ & 16.3273 & \textbf{13.7788} & 15.5958 & 16.2915 \\
$P_1$ & 16.8654 & \textbf{14.1434} & 16.1400 & 16.8353 \\
$\Sigma$ & 18.2635 & 15.1843 & 17.5619 & 18.2470 \\
$\Sigma - P_1$ & 18.5859 & 15.4160 & 17.8875 & 18.5692 \\
\bottomrule
\end{tabularx}
\end{table}

\begin{table}
\centering
\caption{MSE of HR modeling on \textbf{GCDT} using different scaling factors and different data portions.}
\label{tab:gcdt}
\small
\begin{tabularx}{\linewidth}{llllX}
\toprule
 & Baseline & EDU & Sentence & Whole Text \\
\midrule
$P_2$ & 6.8699 & \textbf{6.3902} & 6.8070 & 6.8489 \\
$P_1$ & 6.9967 & \textbf{6.5097} & 6.9303 & 6.9789 \\
$\Sigma$ & 7.2485 & 6.5961 & 7.1511 & 7.2279 \\
$\Sigma - P_1$ & 7.2832 & 6.5644 & 7.1673 & 7.2554 \\
\bottomrule
\end{tabularx}
\end{table}

\begin{figure*}
    \centering
    \includegraphics[width=.9\linewidth]{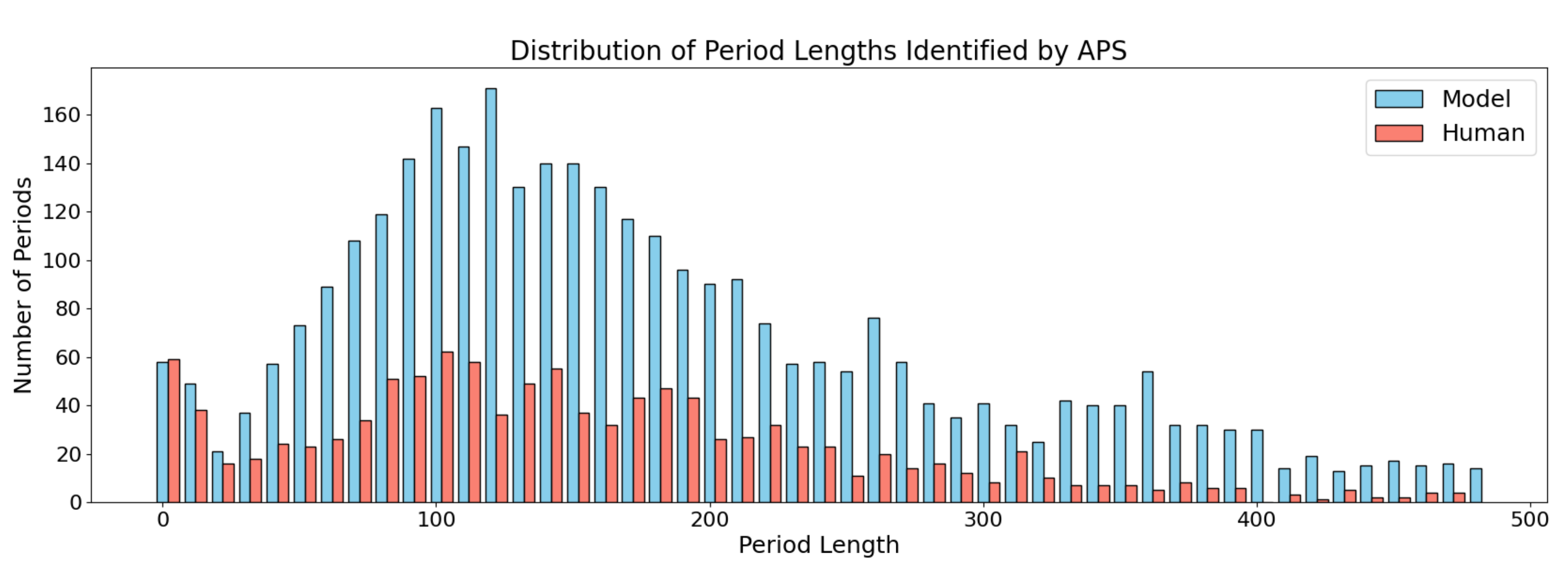}
    \caption{Distributions of valid periods from human (FACE BBC News) and LLM-generated (LLaMA-3-70B) texts.}
    \label{fig:human_vs_generated}
\end{figure*}


\section{Result: Human vs. Generated Texts}
\label{sec:results_human_vs_generated}

By applying \texttt{APS} on the corpora of human-written and LLM-generated text pairs (FACE and EvoBench; see \Cref{sec:corpora}), we can observe periodicity differences between them at the document level. These datasets contain generated texts obtained from LLMs by feeding the beginning of a human-written text as a prompt.
In general, we find that LLM-generated texts contain stronger periodicity of information than human-written ones in both English and Chinese. For instance, the proportion of documents with valid periods in LLaMA3-70B-generated fake BBC news is $30.06\%$, which is more than twice the proportion in the real news ($14.80\%$). 
Detailed results are listed in \Cref{tab:human_vs_generated}. 

\Cref{fig:human_vs_generated} shows the distributions of the detected periods in human-written and LLM-generated texts in FACE BBC News data. It can be observed that the LLM-generated texts exhibit more periodic patterns than human-written texts, especially in the long-period range ($>50$ tokens). 
One of the possible reasons for the stronger periodicity in LLM-generated texts is that LLMs may generate repetitive content. Sometimes, the generated text may contain repeated phrases or sentences, which can create a strong periodic pattern. We randomly sample $10\%$ of the documents with valid periods in LLaMA3-70B-generated BBC news (150 documents in total), and find that $7.33\%$ of them (11~documents) contain repeated phrases or sentences. Thus, repetition could be one of the factors contributing to the difference, especially in the short-period range. However, the gap between the ratios of documents with valid periods in human-written and LLM-generated texts is still significant ($14.80\%$ vs. $27.86\%$) after excluding the repeated portion.
Another possible reason is that LLMs are more likely to generate similar structures across paragraphs, maintaining structural coherence under large spans. Human writers, however, tend to be more flexible in their narrative structures, avoiding orderly patterns. This could lead to stronger periodicity in LLM-generated texts in the long-period range.

The above finding lends interpretation to some existing studies that utilize periodicity of information as a feature for the task of detecting model-generated texts \citep{xu-etal-2024-detecting} and for automatic NLG evaluation \citep{face2-emnlp-2025}. Through \texttt{APS}, we can explicitly tell \emph{how} the periodicity of information in LLM-generated texts is different from that in human-written ones. 
Therefore, information periodicity holds promise as a distinctive feature for generation-related tasks.

\section{Conclusions}

In this study, we propose \texttt{AutoPeriod} of Surprisal (\texttt{APS}), an algorithm for revealing periods of surprisal in natural language with a high resolution, which contributes to the emerging research topic of periodicity of information in human language. Our main observations and conclusions are as follows: 

\emph{Periodicity can be detected}. Our method confirms that, although it is a subtle phenomenon, the periodicity of information can be detected at the single document level. It is validated that there is a considerable proportion of human-written language that exhibits significant periodicity of information. 

\emph{Sources of periodicity}. We find evidence for periods that are not purely due to the linguistic structures, such as sentences, paragraphs, and EDUs. This indicates that there are other factors than structural information processing pressure that cause periodicity, which needs to be studied in future work. We speculate that the longer periods detected could be due to some latent structures (such as topic) that span across multiple paragraphs. 

\emph{Usability in human-machine distinction}. \texttt{APS} is not a regression-based technique that focuses on analysis, but rather a detection method that attempts to extract useful cues about how information distributes temporally given any input text. Therefore, it has potential in distinguishing human- vs. model-generated texts, under the basic assumption that human language production is a cognitively-constrained process.

\section*{Limitations}

Within the scope of this study, we consider the
following limitations, which we believe can be further addressed in future work.
Firstly, this study only examines English and Chinese texts. As languages differ in structure, we believe that our proposed \texttt{APS} method might vary in performance across different texts from different language.

Secondly, our detection results in \Cref{sec:proportions_periodic_docs} suggest that the periodicity of information is influenced by text genre. We also find that parts of the detected periods cannot be explained by local structural units, as discussed in \Cref{sec:comparing_periods_with_units}. However, we do not further analyze the sources of these periods or potential factors that influence the periodicity of information, leaving these inquiries to future work.

Thirdly, the performance of \texttt{AutoPeriod} might not be as accurate in detecting signals with short periods as those with long periods. The corresponding frequencies of the component signals returned by DFT distribute unevenly on the axis of period length. Components with short periods are more densely distributed than those with long periods. When searching for period hints by random perturbation, short periods are thus more likely to be false positives. From the distribution of the detected periods, we observe that the prominent peak in the region of short periods tends to shrink when the confidence level is set higher. The observation indicates that the detection of short periods is less robust.

\section*{Ethics Statements}
\label{sec:ethics}

A potential ethical issue of this research results from the discovery of significant differences in periodicity of information between human- and model-generated texts (\Cref{sec:results_human_vs_generated}). While a promising application of this is the detection of LLM-generated text, such applications have the potential to be misused. Direct conclusions should not be drawn when a text is identified as having unusually high periodicity, as this could be caused by various factors, such as the use of an LLM-based writing assistant, the author's individual style, or the author writing in a second language (which can result in more regular, formulaic expression). Detection results should thus be approached with caution.

The text examples shown in \Cref{fig:period_hints,fig:text-sample} are for illustrative purposes only, which is in compliance with the fair use license of the \href{https://www.ldc.upenn.edu/data-management/using/licensing}{Linguistic Data Consortium}.

\section*{Acknowledgments}

We sincerely thank Eleftheria Tsipidi and Mario Giulianelli for providing the data for harmonic regression analysis and for the insightful discussions. We also thank all the reviewers for their feedback on the paper. 
This study is funded by Shenzhen Science and Technology Program (No. JCYJ20240813094612017) and Guangdong Province ZJRC Program (No. 2024QN11X145).
This study is also funded by the Deutsche For\-schungs\-ge\-mein\-schaft (DFG, German Research Foundation): \href{https://gepris.dfg.de/gepris/projekt/438445824}{TRR 318/3 2026 -- 438445824}, B07.

\bibliography{period2025}

\clearpage
\appendix

\section{Implementation Details of \texttt{APS}}

A full version of the \texttt{APS} algorithm is presented in \Cref{alg:period_acf_full}. In contrast to the abbreviated pseudo-code presented above in \Cref{alg:period_acf}, it details the process of determining the search window $W$, the linear regression-based hill validation, and the final period selection strategy.

\begin{algorithm*}
\caption{ACF Filtering (full version)}
\label{alg:period_acf_full}
\begin{algorithmic}[1]
\Function{ACFFiltering}{$\mathbf{x}$, $\textit{periodHints}$}
    \State $N \gets \mathbf{x}.\textit{length}$
    \State $\textit{ACF} \gets \textsc{getACF}(\mathbf{x})$
    \State $M \gets \textit{periodHints}.\textit{length}$
    \State $\textit{refinedPeriods} \gets \{\}$ 
    
    \For{$i = 1$ to $M$}
        \State $\tau \gets \textit{periodHints}[i]$
        \State $\tau \gets N/k \gets \textit{periodHints}[i]$

        \Comment{Determine search window using adjacent periods}
        \State $\tau_{\text{prev}} \gets N/(k-1)$
        \State $\tau_{\text{next}} \gets N/(k+1)$
        \State $W \gets [\frac{\tau + \tau_{\text{next}}}{2} - 1, \frac{\tau + \tau_{\text{prev}}}{2} + 1]$

        \Comment{Find optimal split point using linear regression}
        \State $\epsilon_{\min} \gets \infty$
        \State $t_{\text{best}} \gets -1$
        \State $\textit{slope}_L \gets 0, \textit{slope}_R \gets 0$
        \For{each $t$ in $W$}
            \State Split $W$ at $t$ into $L$ and $R$
            \State $\textit{slope}^t_L, \epsilon^t_L \gets \textsc{LinearRegression}(L, \textit{ACF}[L])$
            \State $\textit{slope}^t_R, \epsilon^t_R \gets \textsc{LinearRegression}(R, \textit{ACF}[R])$

            \If{$\epsilon^t_L + \epsilon^t_R < \epsilon_{\min}$}
                \State $\epsilon_{\min} \gets \epsilon^t_L + \epsilon^t_R$
                \State $t_{\text{best}} \gets t$
                \State $\textit{slope}_L \gets \textit{slope}^t_L, \textit{slope}_R \gets \textit{slope}^t_R$
            \EndIf
        \EndFor
        
        \Comment{Validate if the split forms a hill}
        \State $\theta_L \gets \arctan(\textit{slope}_L) / (\pi/2)$
        \State $\theta_R \gets \arctan(\textit{slope}_R) / (\pi/2)$
        \State $\Delta\theta \gets |\theta_L - \theta_R|$
        \State $\textit{isHill} \gets (\textit{slope}_L > \textit{slope}_R) \land (\Delta\theta > 0.01)$

        \If{$\textit{isHill}$}
            \Comment{Select peak within window as final period}
            \State $\tau_{\text{refined}} \gets \textsc{findPeak}(\textit{ACF}, W, t_{\text{best}})$
            \State $\textit{refinedPeriods}.\textsc{add}(\tau_{\text{refined}})$
            
        \EndIf
    \EndFor
    \State \Return $\textit{refinedPeriods}$
\EndFunction
\end{algorithmic}
\end{algorithm*}

\section{Amplitudes and Corresponding \textit{p}-Values of HR Modeling}

To verify the periodicity of information identified by \texttt{APS}, we fit HR models using the hints and valid periods returned by \texttt{APS} as the scaling factors, and examine the significance of the harmonic terms. 
We simplify the experimental setting of HR modeling, picking 10 terms ($K=10$) in \Cref{eq:hr_term}, and fitting the model on the entire document set of WSJ. The amplitudes $A_k$ and the corresponding p-values of the coefficients $\beta_{1,k}$ and $\beta_{2,k}$ are reported in \Cref{tab:wsj_coef_full_P1,tab:wsj_coef_full_P2}.

\begin{table}
\centering
\caption{Significance test of all the amplitudes in HR models using \textbf{\texttt{APS} \emph{hints}} as scaling factors on WSJ ($P_1$ portion).}
\label{tab:wsj_coef_full_P1}
\small
\begin{tabularx}{\linewidth}{lXXXX}
\toprule
$k$ & $A_k$ & $\beta$ & coef & $P > |t|$ \\
\midrule
\multirow{2}{*}{1} & \multirow{2}{*}{0.1772}
 & $\beta_{1,1}$ & 0.0736 & 0.000 \\
 & & $\beta_{2,1}$ & 0.1612 & 0.000 \\
\multirow{2}{*}{2} & \multirow{2}{*}{0.0255}
 & $\beta_{1,2}$ & 0.0228 & 0.023 \\
 & & $\beta_{2,2}$ & 0.0115 & 0.227 \\
\multirow{2}{*}{3} & \multirow{2}{*}{0.0539}
 & $\beta_{1,3}$ & 0.0335 & 0.001 \\
 & & $\beta_{2,3}$ & 0.0423 & 0.000 \\
\multirow{2}{*}{4} & \multirow{2}{*}{0.0462}
 & $\beta_{1,4}$ & 0.0246 & 0.016 \\
 & & $\beta_{2,4}$ & 0.0391 & 0.000 \\
\multirow{2}{*}{5} & \multirow{2}{*}{0.0731}
 & $\beta_{1,5}$ & 0.0503 & 0.000 \\
 & & $\beta_{2,5}$ & -0.0531 & 0.000 \\
\multirow{2}{*}{6} & \multirow{2}{*}{0.0373}
 & $\beta_{1,6}$ & 0.0034 & 0.749 \\
 & & $\beta_{2,6}$ & 0.0371 & 0.000 \\
\multirow{2}{*}{7} & \multirow{2}{*}{0.0575}
 & $\beta_{1,7}$ & 0.0036 & 0.723 \\
 & & $\beta_{2,7}$ & -0.0574 & 0.000 \\
\multirow{2}{*}{8} & \multirow{2}{*}{0.0318}
 & $\beta_{1,8}$ & 0.0304 & 0.003 \\
 & & $\beta_{2,8}$ & 0.0093 & 0.326 \\
\multirow{2}{*}{9} & \multirow{2}{*}{0.0191}
 & $\beta_{1,9}$ & 0.0163 & 0.113 \\
 & & $\beta_{2,9}$ & 0.0099 & 0.290 \\
\multirow{2}{*}{10} & \multirow{2}{*}{0.0406}
 & $\beta_{1,10}$ & 0.0343 & 0.001 \\
 & & $\beta_{2,10}$ & -0.0217 & 0.021 \\
\bottomrule
\end{tabularx}
\end{table}

\begin{table}
\centering
\caption{Significance test of all the amplitudes in HR models using \textbf{\texttt{APS} \emph{periods}} as scaling factors on WSJ ($P_2$ portion).}
\label{tab:wsj_coef_full_P2}
\small
\begin{tabularx}{\linewidth}{lXXXX}
\toprule
$k$ & $A_k$ & $\beta$ & coef & $P > |t|$ \\
\midrule
\multirow{2}{*}{1} & \multirow{2}{*}{0.2022}
 & $\beta_{1,1}$ & 0.1273 & 0.000 \\
 & & $\beta_{2,1}$ & 0.1571 & 0.000 \\
\multirow{2}{*}{2} & \multirow{2}{*}{0.0449}
 & $\beta_{1,2}$ & 0.0316 & 0.007 \\
 & & $\beta_{2,2}$ & 0.0319 & 0.005 \\
\multirow{2}{*}{3} & \multirow{2}{*}{0.0662}
 & $\beta_{1,3}$ & 0.0524 & 0.000 \\
 & & $\beta_{2,3}$ & 0.0405 & 0.000 \\
\multirow{2}{*}{4} & \multirow{2}{*}{0.0467}
 & $\beta_{1,4}$ & 0.0367 & 0.002 \\
 & & $\beta_{2,4}$ & 0.0289 & 0.010 \\
\multirow{2}{*}{5} & \multirow{2}{*}{0.0466}
 & $\beta_{1,5}$ & 0.0453 & 0.000 \\
 & & $\beta_{2,5}$ & -0.0111 & 0.318 \\
\multirow{2}{*}{6} & \multirow{2}{*}{0.0296}
 & $\beta_{1,6}$ & 0.0282 & 0.018 \\
 & & $\beta_{2,6}$ & -0.0090 & 0.420 \\
\multirow{2}{*}{7} & \multirow{2}{*}{0.0216}
 & $\beta_{1,7}$ & 0.0190 & 0.108 \\
 & & $\beta_{2,7}$ & -0.0101 & 0.369 \\
\multirow{2}{*}{8} & \multirow{2}{*}{0.0299}
 & $\beta_{1,8}$ & 0.0133 & 0.262 \\
 & & $\beta_{2,8}$ & -0.0268 & 0.018 \\
\multirow{2}{*}{9} & \multirow{2}{*}{0.0310}
 & $\beta_{1,9}$ & 0.0234 & 0.050 \\
 & & $\beta_{2,9}$ & 0.0203 & 0.069 \\
\multirow{2}{*}{10} & \multirow{2}{*}{0.0083}
 & $\beta_{1,10}$ & 0.0036 & 0.762 \\
 & & $\beta_{2,10}$ & -0.0075 & 0.504 \\
\bottomrule
\end{tabularx}
\end{table}

\section{Full Results of Human vs. Generated Texts}

We provide the full results of the detection of periods in human-written vs. LLM-generated texts in \Cref{tab:human_vs_generated}, including the counts and ratios of documents with non-empty period hints ($|P_1|$) and valid periods ($|P_2|$) identified by \texttt{APS} in three corpora. For FACE, we choose the News portion of the dataset in both English and Chinese, and select the texts generated by LLaMA3-70B and Qwen2-72B as the LLM-generated texts, respectively. For EvoBench, we choose the XSum portion of the dataset, and select the subset generated by GPT-4o as the LLM-generated texts.

\begin{table*}[htbp]
\centering
\caption{Counts and ratios of documents with non-empty period hints ($|P_1|$) and valid periods ($|P_2|$) identified by \texttt{APS} in human-written vs. LLM-generated texts across three corpora.}
\begin{tabularx}{\linewidth}{lXXXlXX}
\toprule
 & \multicolumn{2}{c}{$\text{EvoBench}_{\text{en}}$}
 & \multicolumn{2}{c}{$\text{FACE}_{\text{en}}$} 
 & \multicolumn{2}{c}{$\text{FACE}_{\text{zh}}$} \\
\cmidrule(lr){2-3}\cmidrule(lr){4-5}\cmidrule(lr){6-7}
 & Human & GPT-4o & Human & LLaMA3-70B & Human & Qwen2-72B \\
\midrule
$|\Sigma|$ & 150 & 150 & 5000 & 5000 & 5000 & 5000 \\
$|P_1|$ & 9 & 19 & 1032 & 1882 & 1209 & 1592 \\
$|P_2|$ & 5 & 11 & 740 & 1503 & 943 & 1231 \\
\midrule
$\sfrac{|P_1|}{|\Sigma|}$ & 6.00\% & 12.67\% & 20.64\% & 37.64\% & 24.18\% & 31.84\% \\
$\sfrac{|P_2|}{|P_1|}$ & 55.56\% & 57.89\% & 71.71\% & 79.86\% & 78.00\% & 77.32\% \\
$\sfrac{|P_2|}{|\Sigma|}$ & 3.33\% & 7.33\% & 14.80\% & 30.06\% & 18.86\% & 24.62\% \\
\bottomrule
\end{tabularx}
\label{tab:human_vs_generated}
\end{table*}

\section{A Complete Example of \texttt{APS} on a WSJ Document}

\begin{figure*}
    \centering
    \begin{subfigure}[b]{0.32\textwidth}
        \centering
        \includegraphics[width=\textwidth]{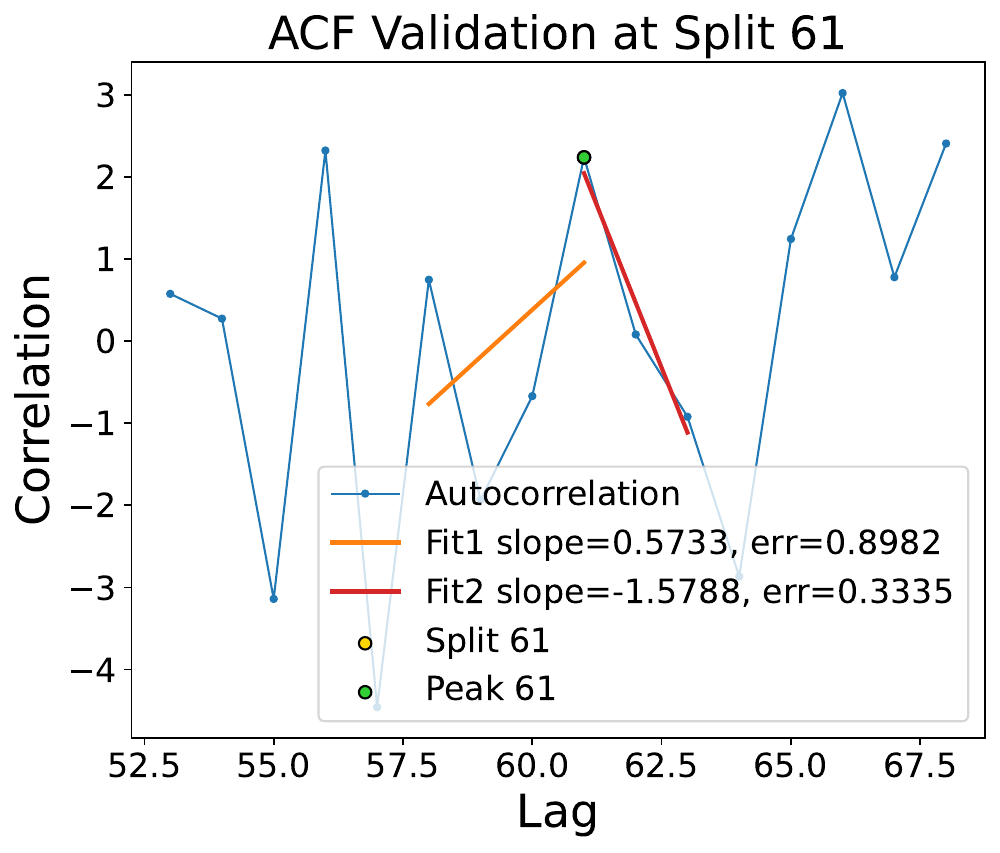}
        \caption{ACF at the period hint of 61 tokens.}
        \label{fig:example_acf_sub_a}
    \end{subfigure}
    \hfill 
    \begin{subfigure}[b]{0.32\textwidth}
        \centering
        \includegraphics[width=\textwidth]{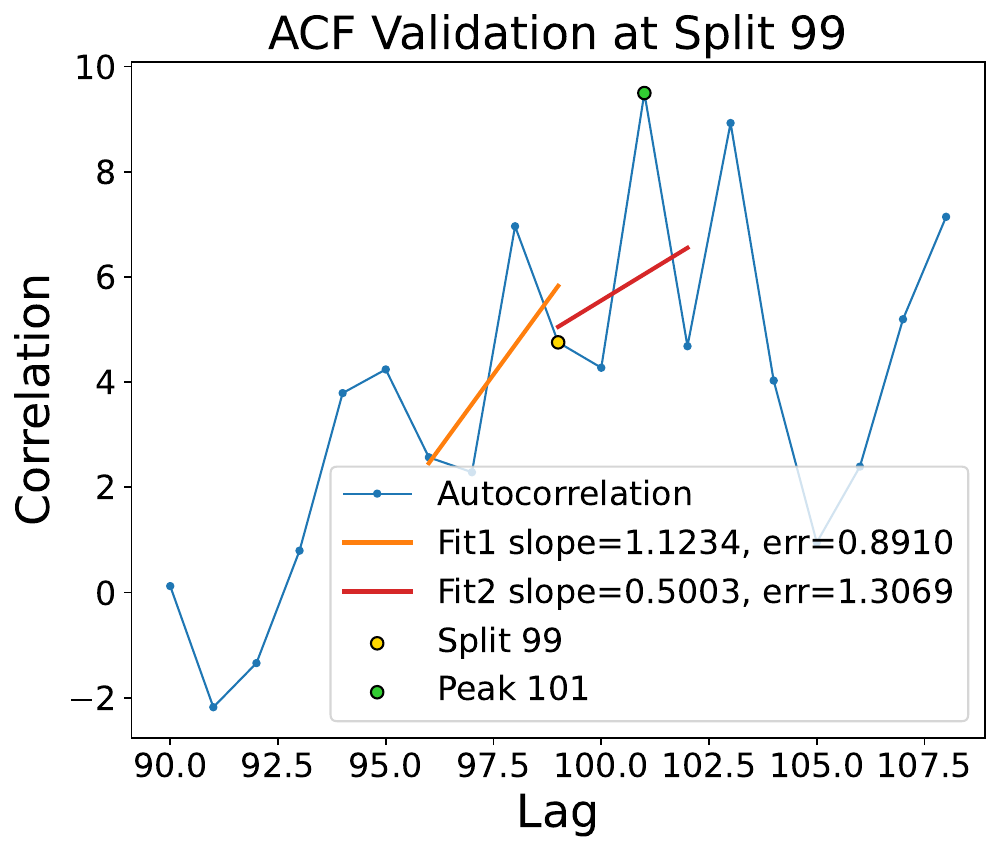}
        \caption{ACF at the period hint of 99 tokens.}
        \label{fig:example_acf_sub_b}
    \end{subfigure}
    \hfill
    \begin{subfigure}[b]{0.32\textwidth}
        \centering
        \includegraphics[width=\textwidth]{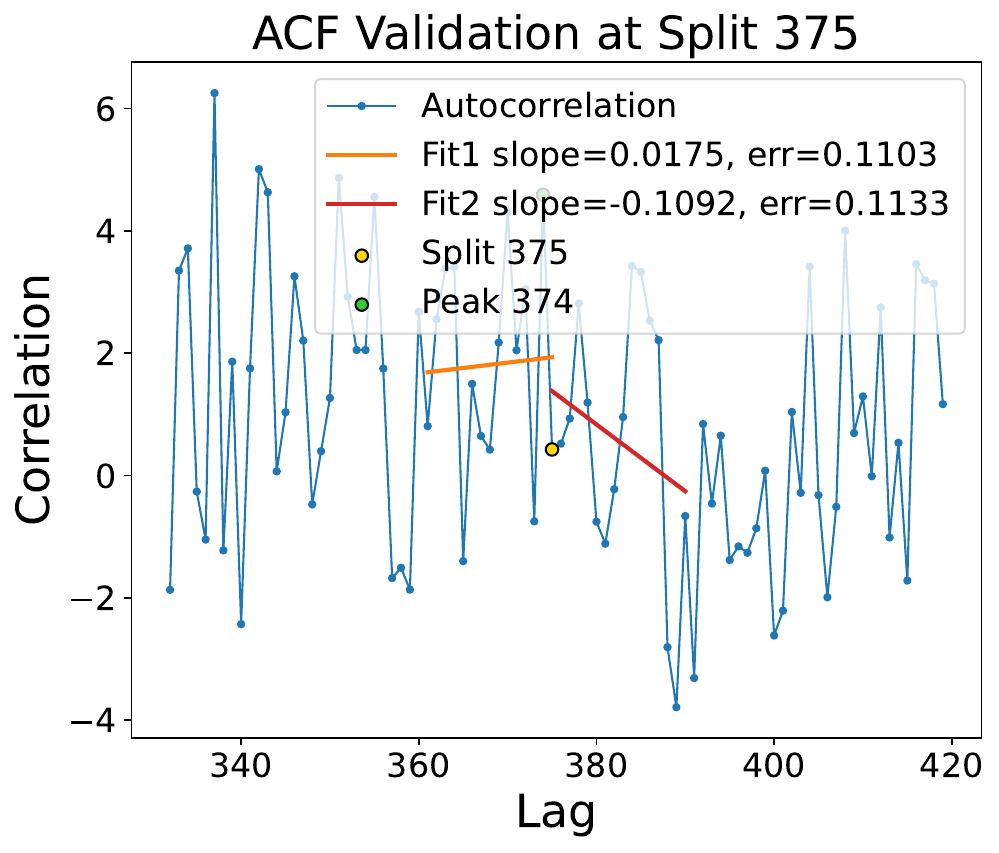}
        \caption{ACF at the period hint of 375 tokens.}
        \label{fig:example_acf_sub_c}
    \end{subfigure}

    \caption{Examples of the ACF curves at the period hints 61, 99, and 375 returned by \texttt{APS}. The yellow split points are the optimal split points determined by the linear regression-based hill validation, the green peak points are the peaks within the search windows.}
    \label{fig:example_acf}
\end{figure*}

For a text sample from document \#2380 from PTB-WSJ (presented in \Cref{fig:text-sample}), we apply \texttt{APS} to detect the periods of surprisal. In the token surprisal curve, shown in \Cref{fig:example_timeseries}, we mark the intervals of paragraphs with dashed red lines. The average length of the paragraphs is 61.76 tokens. The periodogram of the surprisal curve is shown in \Cref{fig:example_periodogram}, where the dashed lines represent different confidence levels, and the period hints above the 99\% confidence level are marked with crosses. The candidate periods are 60, 99, 101, 374, and 403 tokens. The ACF curves at the period hints of 61, 99, and 375 tokens (the best splits within the windows) are shown in \Cref{fig:example_acf_sub_a}, \Cref{fig:example_acf_sub_b} and \Cref{fig:example_acf_sub_c} respectively. 
We use ACF curves to filter and refine the period hints. For this example, the period hint of 403 tokens is filtered out, and the period hint of 99 tokens is refined to 101 tokens.
The final detected valid periods are 61, 101, and 374 tokens.

\begin{figure*}
    \centering
    \begin{tcolorbox}[
        width=\linewidth,
        colback=white,
        colframe=black,
        sharp corners,
        boxrule=0.5pt]

\ttfamily
Friday , October 13 , 1989

The key U.S. and foreign annual interest rates below are a guide to general levels but do n't always represent actual transactions .

PRIME RATE : 10 1/2 \% . The base rate on corporate loans at large U.S. money center commercial banks .

FEDERAL FUNDS : 8 13/16 \% high , 8 1/2 \% low , 8 5/8 \% near closing bid , 8 3/4 \% offered . Reserves traded among commercial banks for overnight use in amounts of \$ 1 million or more . Source : Fulton Prebon ( U.S.A . ) Inc .

DISCOUNT RATE : 7 \% . The charge on loans to depository institutions by the New York Federal Reserve Bank .

...

FEDERAL HOME LOAN MORTGAGE CORP . ( Freddie Mac ) : Posted yields on 30-year mortgage commitments for delivery within 30 days . 9.91 \% , standard conventional fixedrate mortgages ; 7.875 \% , 2 \% rate capped one-year adjustable rate mortgages . Source : Telerate Systems Inc .

FEDERAL NATIONAL MORTGAGE ASSOCIATION ( Fannie Mae ) : Posted yields on 30 year mortgage commitments for delivery within 30 days ( priced at par ) 9.86 \% , standard conventional fixed-rate mortgages ; 8.85 \% , 6/2 rate capped one-year adjustable rate mortgages . Source : Telerate Systems Inc .

MERRILL LYNCH READY ASSETS TRUST : 8.33 \% . Annualized average rate of return after expenses for the past 30 days ; not a forecast of future returns .
\end{tcolorbox}
\caption{Sample text from document \#2380 from the PTB-WSJ corpus.}
\label{fig:text-sample}
\end{figure*}

\begin{figure*}
    \centering
    \includegraphics[width=\linewidth]{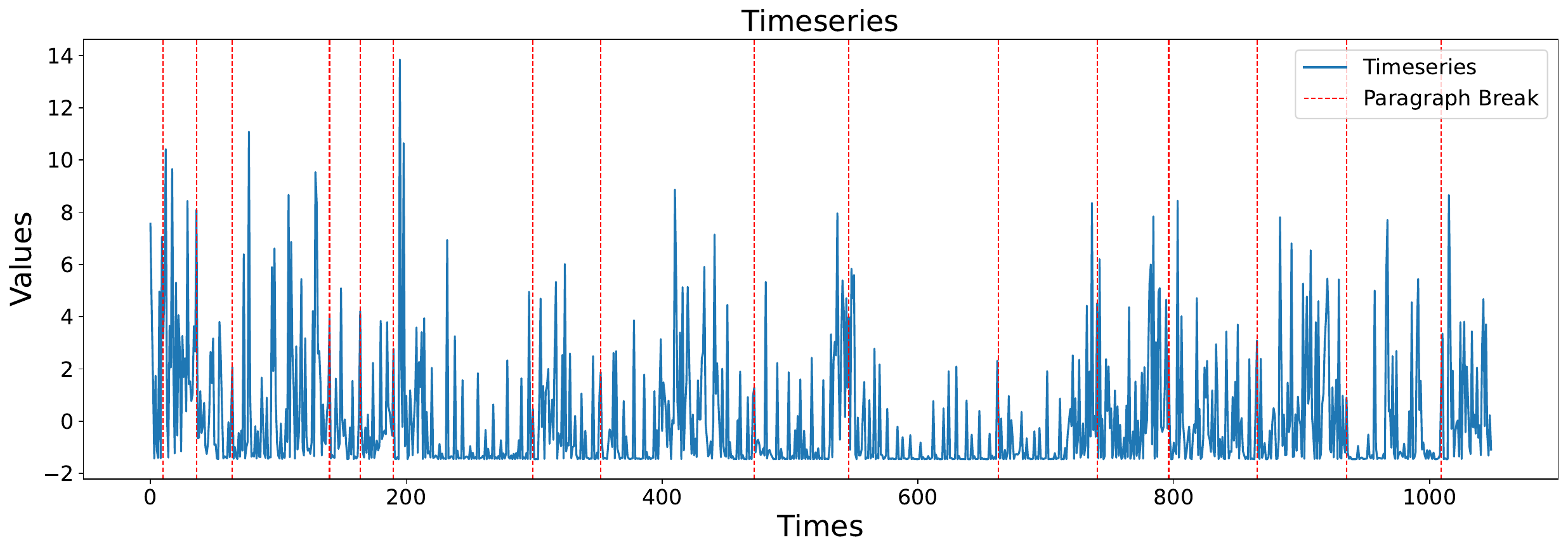}
    \caption{The token surprisal curve of document the \#2380 from the PTB-WSJ corpus shown in \Cref{fig:text-sample}. The dashed red lines indicate the intervals of paragraphs. The average length of the paragraphs is 61.76 tokens.}
    \label{fig:example_timeseries}
\end{figure*}

\begin{figure*}[ht]
    \centering
    \includegraphics[width=\linewidth]{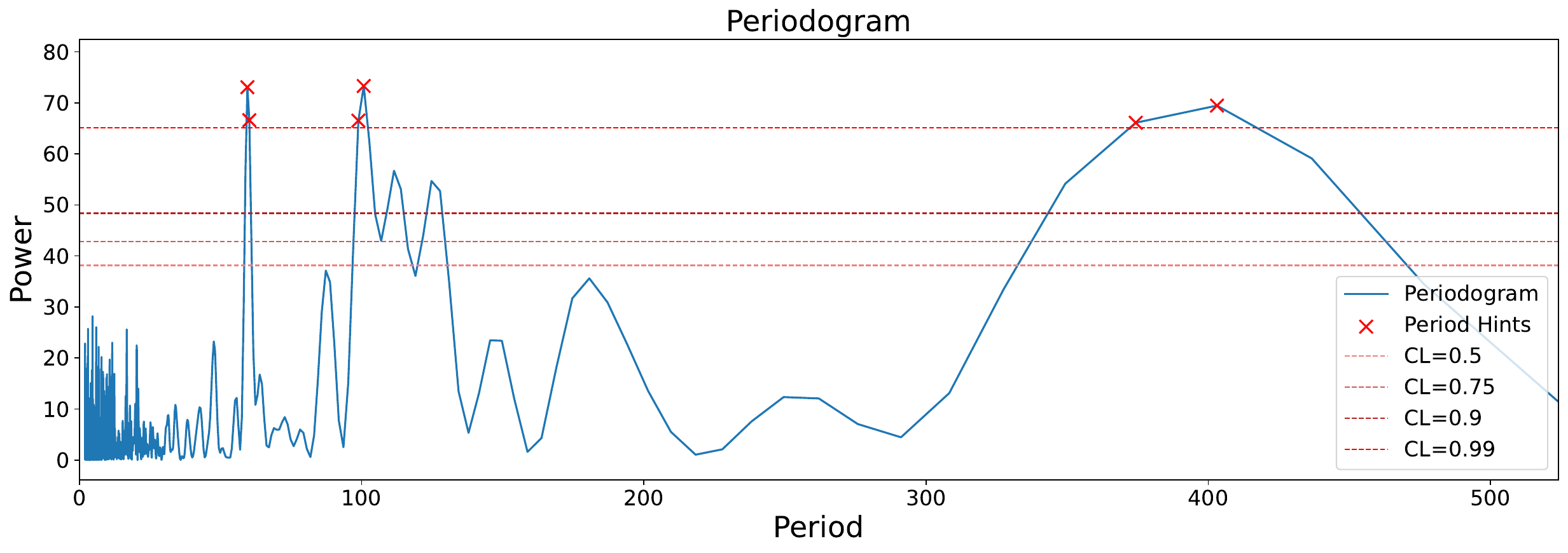}
    \caption{The periodogram of the surprisal curve in \Cref{fig:example_timeseries}. The dashed lines represent different confidence levels. The period hints above the 99\% confidence level are marked with crosses.}
    \label{fig:example_periodogram}
\end{figure*}

\end{document}